\documentclass[10pt,a4paper]{article}
\usepackage{graphicx}
\usepackage{subfigure}
\usepackage{amsmath,amsthm}
\usepackage{amssymb,amsfonts}
\usepackage{authblk}
\usepackage{url}
\usepackage{hyperref}
\usepackage{textgreek}

\newtheorem{definition}{\textbf{Definition}}
\newtheorem{theorem}{\textbf{Theorem}}
\newtheorem{corollary}{\textbf{Corollary}}

\title{Identifying Time Lag in Dynamical Systems with Copula Entropy based Transfer Entropy}
\author{Jian MA\thanks{Email: majian@hitachi.cn}}
\affil{Hitachi China Research Laboratory}
\date{}

\begin{document}

\maketitle

\begin{abstract}
	\noindent
	Time lag between variables is a key characteristics of dynamical systems in different fields and identifying such time lag is an important problem in complex systems with many applications. Transfer Entropy (TE) was proposed as a tool for time lag identification recently. Unfortunately, estimating TE has been a notoriously difficult problem. Copula Entropy (CE) is a measure of statistical independence and it was proved that TE can be represented with only CE. Therefore, a non-parametric estimator of TE based on CE was proposed according to such representation recently. In this paper we propose to use the CE-based estimator of TE to identify time lag in dynamical systems. Both simulated and real data are used to verify the effectiveness of the proposed method in the experiments. Experimental results show that the proposed method can identify the time lags in the four simulated systems. The real data experiment with the data on power consumption of the Tetouan city also demonstrates that our method can identify the pattern of time lags through the estimated TE from the weather factors to the power consumption of the city.
\end{abstract}

\section{Introduction}
The world runs with time lags. In every physical, social, or biological systems, it takes time to see one causal factor's effect on others due to the transmission of mass, energy, or information. Time lag between system variables is one of the key characteristics of dynamical systems in different fields. Identifying such time lags is an important problem in complex systems \cite{Bjorklund2003} and can help us understand causal relationships in the underlying systems. 

One of the traditional intuitive methods for time lag identification is based on temporal correlation between time series of temporal variables \cite{Knapp1976,Gu2011,Kiviet2014}. However, correlation coefficient can only be applied with Gaussian assumptions, which is unrealistic in nonlinear systems with non-Gaussianity. An alternative to temporal correlation is time-delayed mutual information (MI) which can deal with temporal nonlinear correlations \cite{Mars1982,Wen2009,Albers2012,Wilmer2012}. MI is an information theoretical concept that can measure nonlinear statistical dependence between random variables and can be considered as nonlinear generalization of correlations \cite{infobook}. However, both correlation and MI are symmetric and defined for measuring undirectional relationships and therefore is inappropriate for identifying time lag in dynamical systems because the causal relationships associated with time lags are directional.

Transfer Entropy (TE) is another information theoretical concept defined for measuring causal relationships between temporal variables \cite{Schreiber2000}. It is defined as a Kullback-Leibler divergence between two conditional entropies with and without conditioning on causal factor. Essentially, TE is conditional MI and can be considered as nonlinear generalization of Granger Causality (GC) \cite{Granger1969,Granger1980}. It can measure directional information in dynamical systems and was recently proposed as a tool for time lag identification \cite{Lahiri2017,Rosinberg2018,Marzbanrad2015,Oh2023}.

Estimation of TE is of fundamental importance for its real applications and was once considered as a notoriously difficult problem. A basic method for estimating TE is based on the definition of TE, such as \cite{faes2013compensated,Vicente2011}. Such proposed estimators usually present inconsistent and biased estimation results because the key step of estimating densities from limited data used in those estimators are unstable. 

Copula Entropy (CE) is a mathematical concept defined for measuring statistical independence and proved to be equivalent to MI \cite{Ma2011}. Ma proved that TE can be represented with only CE and proposed a nonparametric estimator of TE based on such representation \cite{ma2021estimating}. Since this estimator is rigorous by a sound proof, its performance is consistent and reliable. It has been applied to many real cases, such as fault diagnosis \cite{Liu2022}, sentiment analysis \cite{Han2022}, electricity distribution analysis \cite{Hu2022}.

In this paper, we propose to identify time lag of dynamical systems with TE estimated with the CE-based method. Since the CE-based estimator of TE is nonparametric and therefore can be applied to any system without any assumptions. We will verify the effectiveness of the proposed method with both simulated and real data. In simulation experiments, a group of time series data will be simulated from four different stochastic dynamical systems with different time lags. A real data on energy consumption forecasting of the Tetuan city will be used in our experiments to show the time lag between weather factors and energy consumption identified by our method.

This paper is organized as follows: Section \ref{sec2} introduces some related work, Section \ref{sec3} presents the theory of CE and the methods for estimating CE and estimating TE via CE, Section \ref{sec4} proposes our method for identifying time lag with CE-based TE, Section \ref{sec5} and \ref{sec6} are about the experiments with simulated and real data respectively, Section \ref{sec7} concludes the paper.

\section{Related Work}
\label{sec2}
TE was recently proposed as a tool for identifying time lag in different systems. Lahiri et al. \cite{Lahiri2017} studied the directional influence between cellular processes with TE based time lag identification. The TE in the study was derived with both analytical and numerical methods. Rosinbery et al. \cite{Rosinberg2018} applied TE to stationary linear stochastic processes with a delayed coupling, in which the Wiener-Hoef factorization was adopted for the calculation of TE with respect to time lag. Amornbunchornvej et al. \cite{Amornbunchornvej2021} proposed a method for inferring variable-delay with GC and TE. They demonstrated their method with real applications.

There are several related real applications with TE. Time lag identification with TE was applied to traffic analysis to answer the basic questions on the cause and propagation of traffic jam \cite{Oh2023}, in which the Effective TE method was used to estimate TE along with binning. Marzbanrad et al. \cite{Marzbanrad2015} used TE to study the biological interaction between maternal and fetal heart rates with different time lags, in which TE was computed by binning-based definition. The lag in the information flow between mother and fetus was studied. Wing et al. \cite{Wing2016} investigated the solar wind-earth's radiation belt coupling with information theoretical tools, including MI and TE. They found that TE can be used to detect the change of such coupling and that the time lag from solar wind to geosynchronous MeV electron flux peaks is about 2 days. In their study, TE was also estimated with the binning technique where the number of bins needs to be chosen carefully. 

Estimating TE from the original definition of it was also studied previously. Faes et al. \cite{faes2013compensated} proposed to estimate TE as the difference between two conditional entropies in its definition. Vicente et al. \cite{Vicente2011} proposed to estimate TE as the sum of four entropies expanded from the original definition of TE. Staniek and Lehnertz \cite{Staniek2008} suggested to estimate TE with a technique of symbolization that first rearrange the original variables into symbols and then use symbol-based definition of TE as the estimation result.

Copula was also applied to estimate time delay in dynamical systems. Wang et al. \cite{Wang2017} proposed to use copula-based dependence measure to identify the time delay of linear systems. The measure used in their paper is defined as the distance between empirical copula and product copula.

\section{Methodology}
\label{sec3}
\subsection{Theory of CE}
Copula theory unifies representation of multivariate dependence \cite{nelsen2007,joe2014}. According to Sklar theorem \cite{sklar1959fonctions}, a multivariate joint density function can be represented as a product of its marginals and copula density which represents dependence structure among random variables. Please refer to \cite{Ma2011} for notations.

With copula density, Ma and Sun \cite{Ma2011} defined a new mathematical concept, called Copula Entropy, as follows:
\begin{definition}[Copula Entropy]
	Let $\mathbf{X}$ be random variables with marginals $\mathbf{u}$ and copula density $c(\mathbf{u})$. The CE of $\mathbf{X}$ is defined as
	\begin{equation}
		H_c(\mathbf{x})=-\int_{u}{c(\mathbf{u})\log c(\mathbf{u})d\mathbf{u}}.
	\end{equation}
\end{definition}
In information theory, MI is a fundamental concept different from entropy \cite{infobook}. Ma and Sun \cite{Ma2011} proved that MI is essentially negative CE, as follows:
\begin{theorem}
	MI of random variables is equivalent to negative CE:
	\begin{equation}
		I(\mathbf{x})=-H_c(\mathbf{x}),
	\end{equation}
	where $I$ denotes MI.
	\label{theorem1}
\end{theorem}
Theorem \ref{theorem1} has a simple proof \cite{Ma2011} and an instant corollary on the relationship between the information contained in joint density function, marginals, and copula density.
\begin{corollary}
	The joint entropy of random variables equals to the sum of the entropies of each variable and the CE of the random variables:
	\begin{equation}
		H(\mathbf{x})=\sum_{i}{H_i(x_i)}+H_c(\mathbf{x}),	
	\end{equation}
	where $H$ denotes entropy.
\end{corollary}
The above worthy-a-thousand-word results cast insight into the relationship between MI and copula and therefore build a bridge between information theory and copula theory.

\subsection{Estimating CE}
MI, as a fundamental concept in information theory, has wide applications in different fields. However, estimating it has been notoriously difficult. Under the blessing of Theorem \ref{theorem1}, Ma and Sun \cite{Ma2011} proposed a simple and elegant nonparametric method for estimating CE(MI) from data, which composed of two steps:
\begin{enumerate}
	\item estimating Empirical Copula Density (ECD);
	\item estimating CE from the estimated ECD. 
\end{enumerate}
Given a sample $\{X_1,\ldots,X_T\}$ i.i.d. generated from random variables $\mathbf{X}=\{X_1,\ldots,X_N\}$, one can easily derive ECD using empirical functions:
\begin{equation}
	F_i(x_i)=\frac{1}{T}\sum_{t=1}^{T}{I(X_t^i\leq x_t^i)},
\end{equation}
where $i=1,\ldots,N$ and $I$ denotes indicator function. Once ECD is estimated, estimating CE becomes a problem of entropy estimation which can be tackled with many existing methods. The k-Nearest Neighbor method \cite{Kraskov2004} was suggested in \cite{Ma2011} for such entropy estimation. In this way, a nonparametric method for estimating CE is derived.

\subsection{Estimating TE via CE}
\label{sec:estte}
TE is a information theoretical concept for measuring causal relationships between time series. It is defined from the Wiener's principle which states that if a random variable $X$ can improve the prediction of $Y$ then $X$ is a cause of $Y$. GC is a measure defined from this principle with linear VAR models. TE can be considered as a nonlinear generalization of GC. TE is a model-free measure without any assumptions.

Given two time series $X_i,Y_i,i=1,\ldots,T$, the TE from $X$ to $Y$ is defined as
\begin{equation}
	T_{X\rightarrow Y}=\sum{p(y_{i+1},y^i,x_i)\log\frac{p(y_{i+1}|y^i,x_i)}{p(y_{i+1}|y^i)}},
	\label{eq:te}
\end{equation}
where $p$ denotes probability density function and $y^i=(y_1,\ldots,y_i)$.

Ma \cite{ma2021estimating} proved that TE can be represented with only CE, as follows:
\begin{equation}
	T_{X\rightarrow Y}=-H_c(y_{i+1},y^i,x_i)+H_c(y_{i+1},y^i)+H_c(y_{i+1},x_i)-H_c(y^i).
	\label{eq:tehc}
\end{equation}
based on this representation, Ma \cite{ma2021estimating} proposed a nonparametric estimator of TE which composed of two simple steps:
\begin{enumerate}
	\item estimating the CE terms in \ref{eq:tehc} with the nonparametric CE estimator;
	\item calculating TE from the estimated CE terms.
\end{enumerate}

\section{Proposed Method}
\label{sec4}
In this section we propose to use our TE estimator for identifying time lag in dynamical systems. TE will be used to quantify causal relationship between variables in systems from their time series observations. First, a group of TE between variables will be estimated from time series data by changing the time lag between two variables; and then the lag associated with maximal TE value will be identified as the lag between variables in dynamical systems. TE will be estimated with the non-parametric estimator in Section \ref{sec:estte}, which is model-free and therefore makes no assumption on the underlying systems. The identified time lag will be considered as the characteristics of system dynamics and the associated TE value will be considered as the causal strength between two variables.

\section{Simulations}
\label{sec5}
\subsection{Experiments}
We conducted simulation experiments to verify the effectiveness of the proposed method. In the simulation, four dynamical systems are designed to simulate different dynamics with time lag. Both state delay and output delay are considered in these systems.

\paragraph{System 1}The first system is with a state variable  governed by Gaussian distribution and a output variable which is derived by shifted the state variable with a pre-specified time lag. The mathematical model of the system is represented as follows:
\begin{equation}
\begin{aligned}
X_i &= \xi_1,\\
Y_{i+l} &= X_i + \xi_2, \\
\xi_1&\sim N(\mu_1,\delta_1),\\
\xi_2&\sim N(\mu_2,\delta_2),
\end{aligned}
\end{equation}
where $l$ denotes time lag from the state variable $X$ to the output variable $Y$, and $N(\mu_i,\delta_i)$ denote normal distribution with mean $\mu_i$ and variance $\delta_i$.

\paragraph{System 2}The second system is similar to the first one but with an additional sine dynamical term in the state variable. The mathematical model of the second system is represented as follows:
\begin{equation}
\begin{aligned}
X_i &= \sin(2\pi i/m) + \xi_1,\\
Y_{i+l} &= X_i + \xi_2, \\
\xi_1&\sim N(\mu_1,\delta_1),\\
\xi_2&\sim N(\mu_2,\delta_2),
\end{aligned}
\end{equation}
where $l$ denotes time lag from the state variable $X$ to the output variable $Y$, $N(\mu_i,\delta_i)$ denote normal distribution with mean $\mu_i$ and variance $\delta_i$, and $m$ denotes the length of the simulated time series. Due to the sine term, the dynamical characteristics of the second system is different from that of the first one.

\paragraph{System 3}The third system is first order with a state variable and a output variable with respect to time lag. The mathematical model of the third system is represented as follows:
\begin{equation}
\begin{aligned}
X_i &= X_{i-1} + \xi_1,\\
Y_{i+l} &= X_i + \xi_2, \\
\xi_1&\sim N(\mu_1,\delta_1),\\
\xi_2&\sim N(\mu_2,\delta_2),
\end{aligned}
\end{equation}
where $l$ denotes time lag from the state variable $X$ to the output variable $Y$, $N(\mu_i,\delta_i)$ denote normal distribution with mean $\mu_i$ and variance $\delta_i$. This is actually Wiener process, in which autoregression is introduced on the state variable $X$.

\paragraph{System 4}In the previous three systems, the output-delay is simulated. The fourth system here is first order delayed system simulating delays in state variable. The mathematical model of it is as follows:
\begin{equation}
\begin{aligned}
X_i &= \alpha X_{i-1} + \beta X_{i-l} + \xi_1,\\
\xi_1&\sim N(\mu_1,\delta_1),
\end{aligned}
\end{equation}
where $l$ denotes time lag in system state, $\alpha,\beta$ are the parameters of the system, and $\xi_1$ is a random variable govened by normal distribution with mean $\mu_1$ and variance $\delta_1$. In this setting, the current state of the system is related to not only the last state but also a state of $l$ time lags before.

In these simulation experiments, we will change the value of the time lag $l$ from 1 to 4 to generate 4 trajectories of the state or output variables of each system. Our method will be applied to the generated trajectory data to estimate 1) the TE from state variable to output variable of the first three systems and 2) the TE within state variable of the fourth system with respect to time lags, and then identify the time lag of the system through the maximum of the estimated TE values. We estimated the TE of the time lag from 1 to 8, which is more than the time lag horizon of the simulated system. In all the simulations, $\mu_i=0$, $\delta_i=0.001$, and the length of the simulated time series data is 500. The parameters $\alpha,\beta$ in the fourth system are 0.2 and 0.8 respectively.

\subsection{Results}
\begin{figure}
	\centering
	\subfigure[System 1]{
	\includegraphics[width=0.8\textwidth]{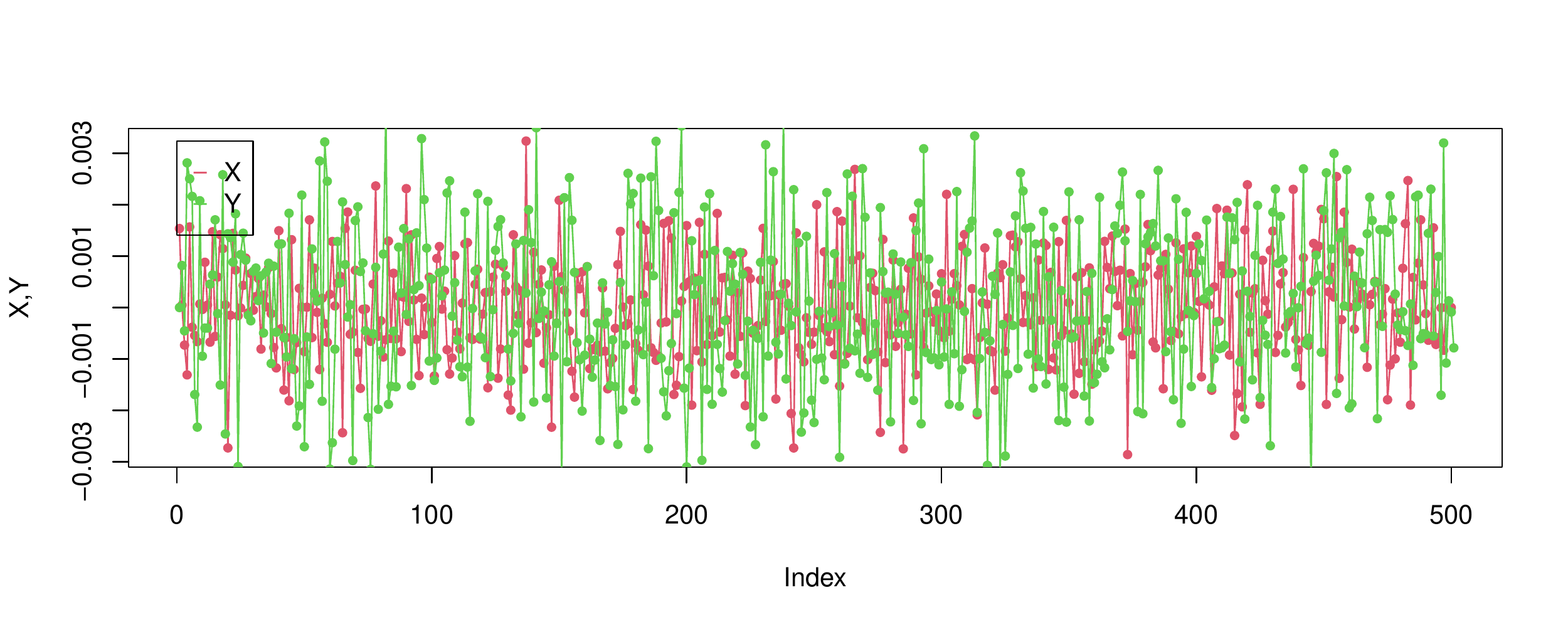}
	\label{fig:sim1data1}}
	\subfigure[System 2]{
	\includegraphics[width=0.8\textwidth]{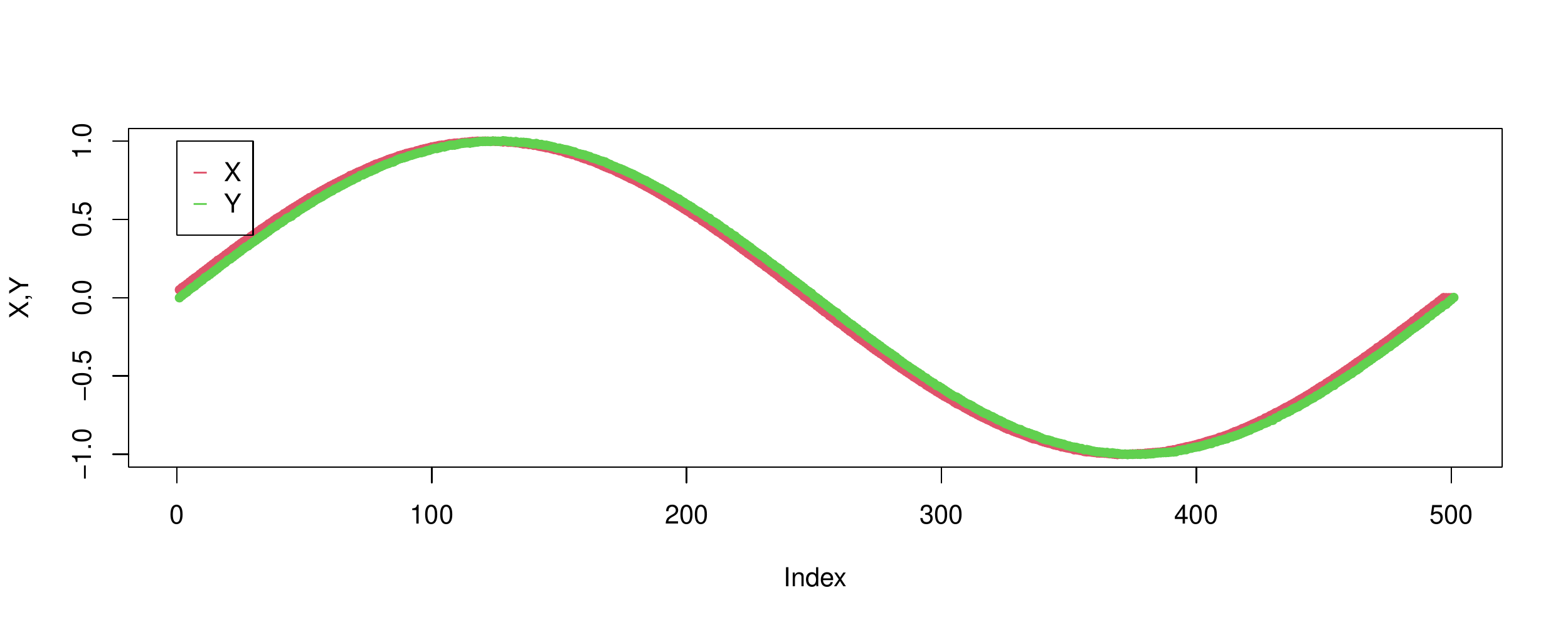}
	\label{fig:sim1data2}}
	\subfigure[System 3]{
	\includegraphics[width=0.8\textwidth]{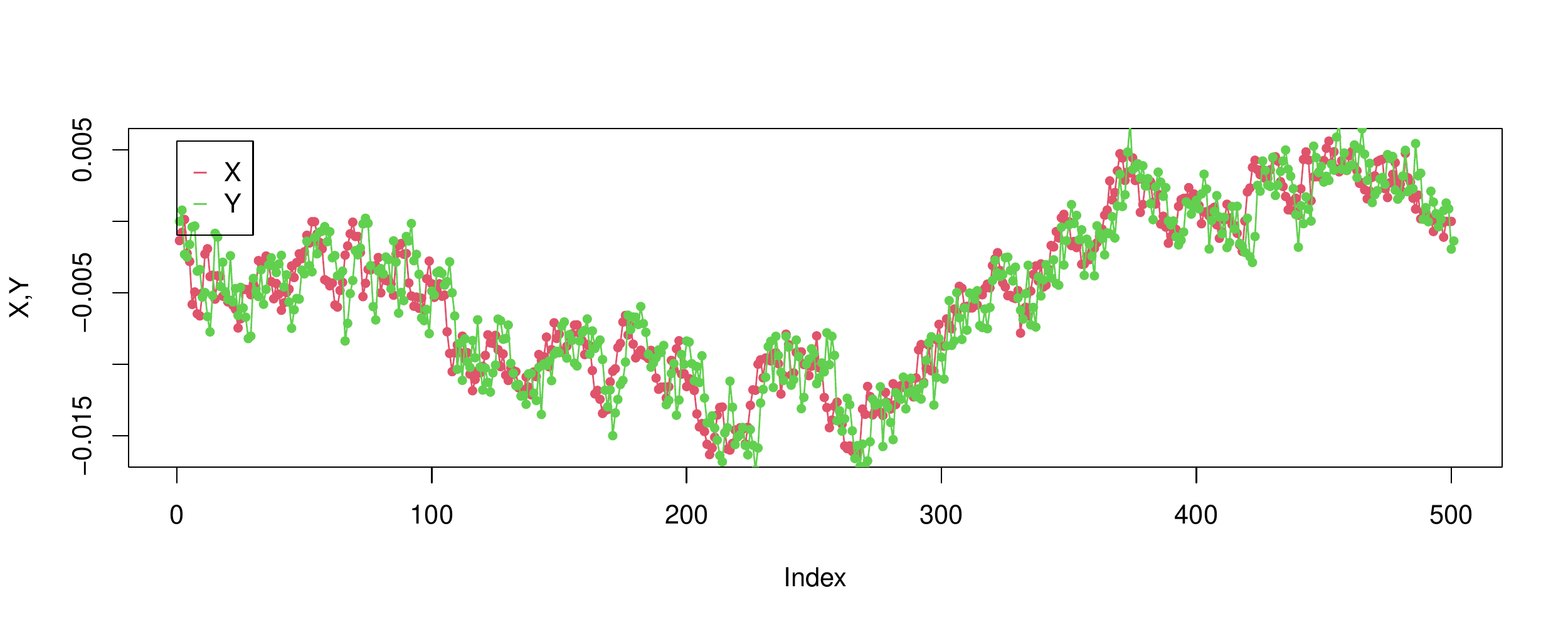}
	\label{fig:sim1data3}}
	\subfigure[System 4]{
	\includegraphics[width=0.8\textwidth]{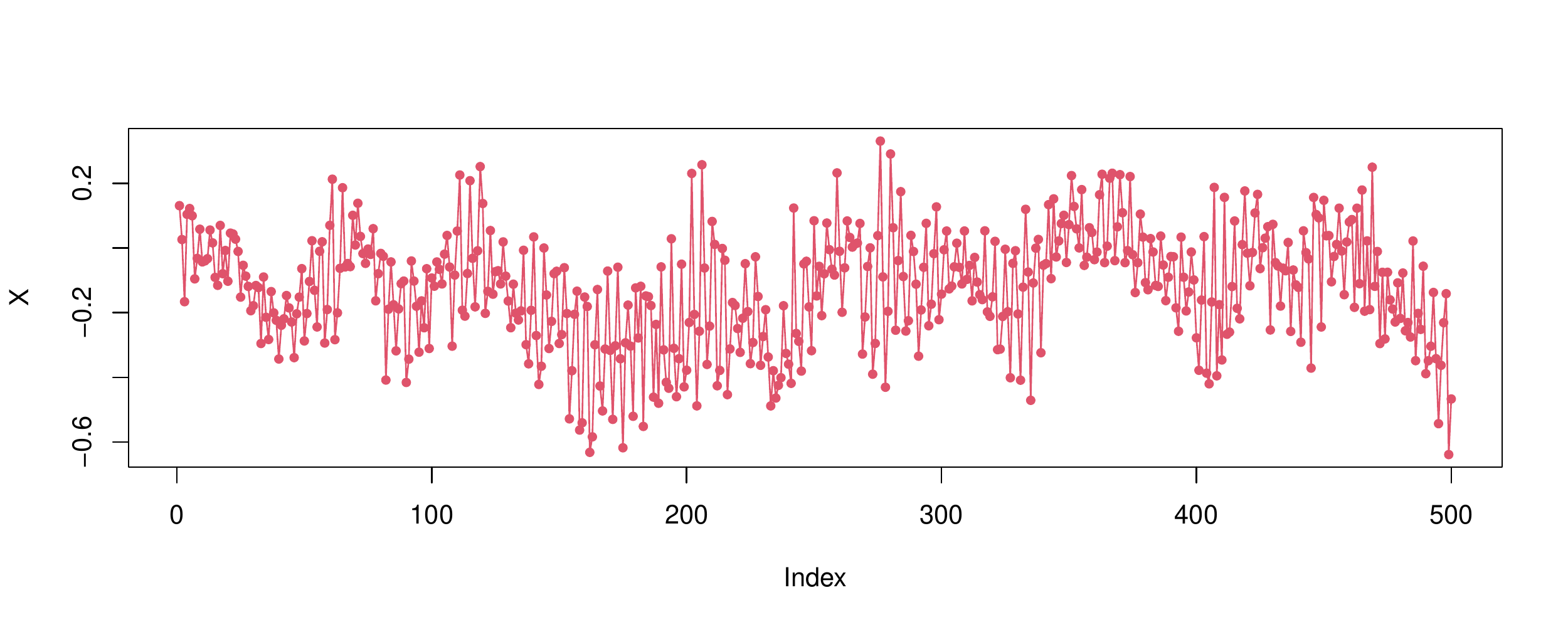}
	\label{fig:sim2}}	
	\caption{Simulated trajectories in the simulation experiments.}
	\label{fig:sim}
\end{figure}

\begin{figure}
	\centering
	\includegraphics[width=0.75\textwidth]{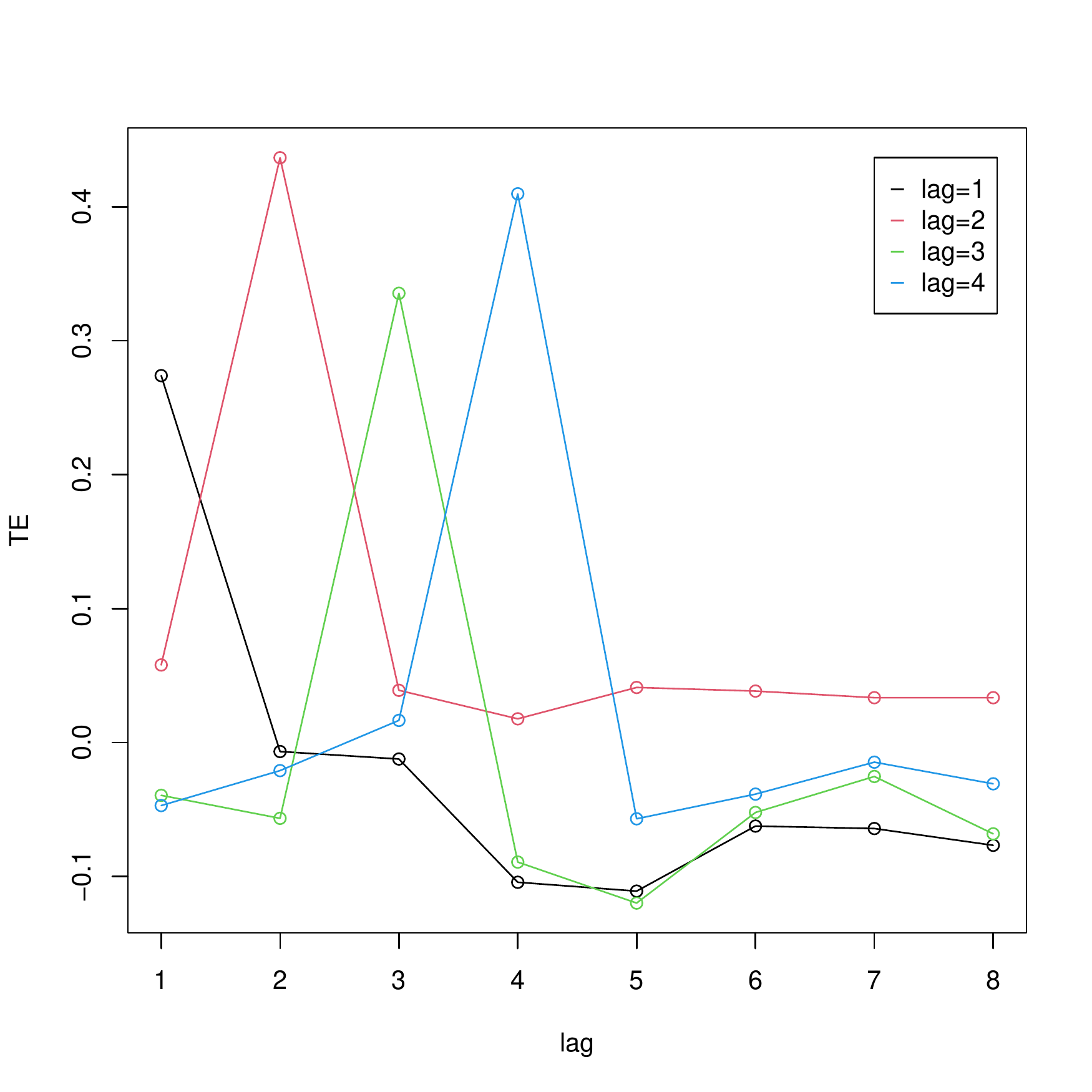}
	\caption{Results of the simulation experiment 1.}
	\label{fig:sim1result1}
\end{figure}

\begin{figure}
	\centering
	\includegraphics[width=0.75\textwidth]{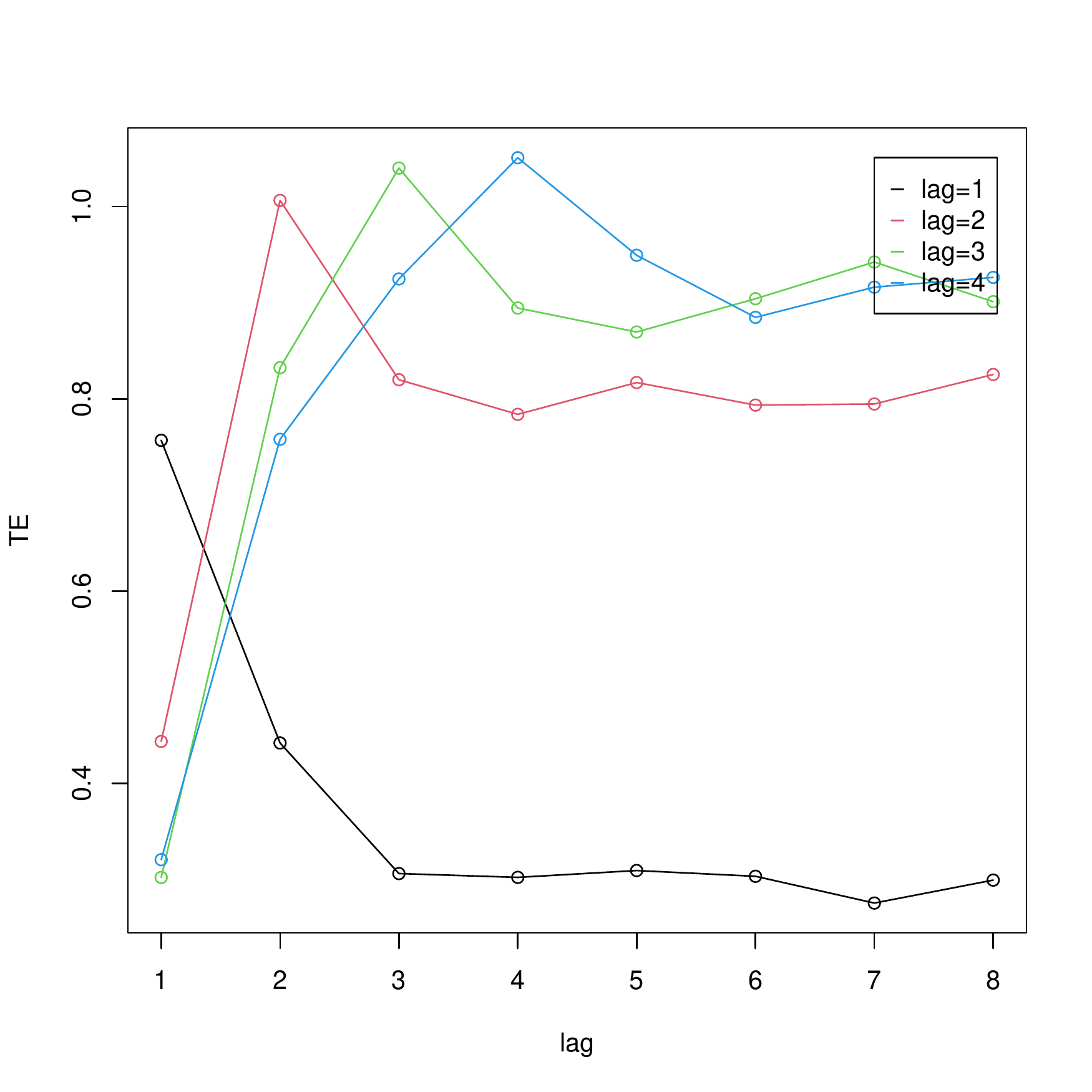}
	\caption{Results of the simulation experiment 2.}
	\label{fig:sim1result2}
\end{figure}

\begin{figure}
	\centering
	\includegraphics[width=0.75\textwidth]{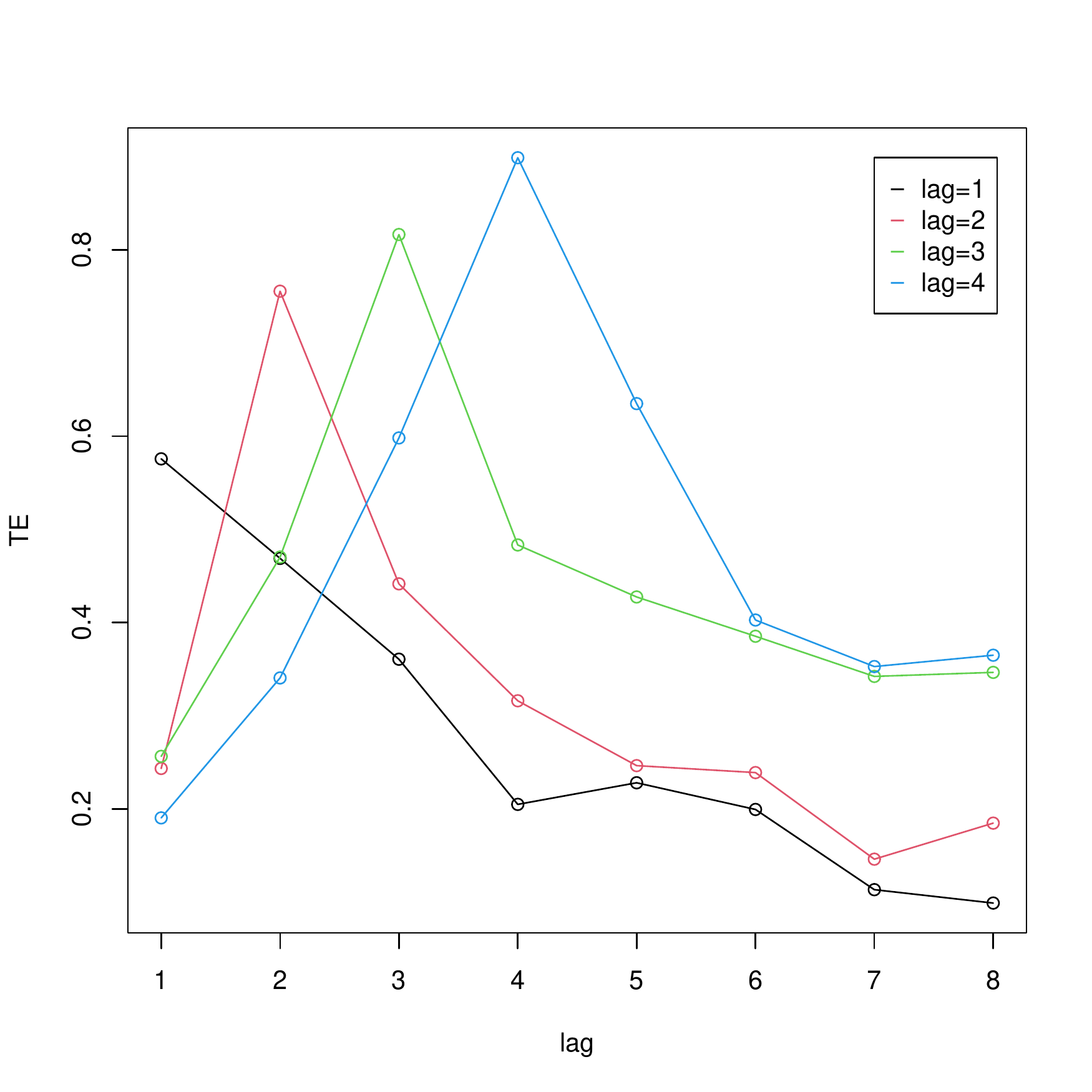}
	\caption{Results of the simulation experiment 3.}
	\label{fig:sim1result3}
\end{figure}

\begin{figure}
	\centering
	\includegraphics[width=0.75\textwidth]{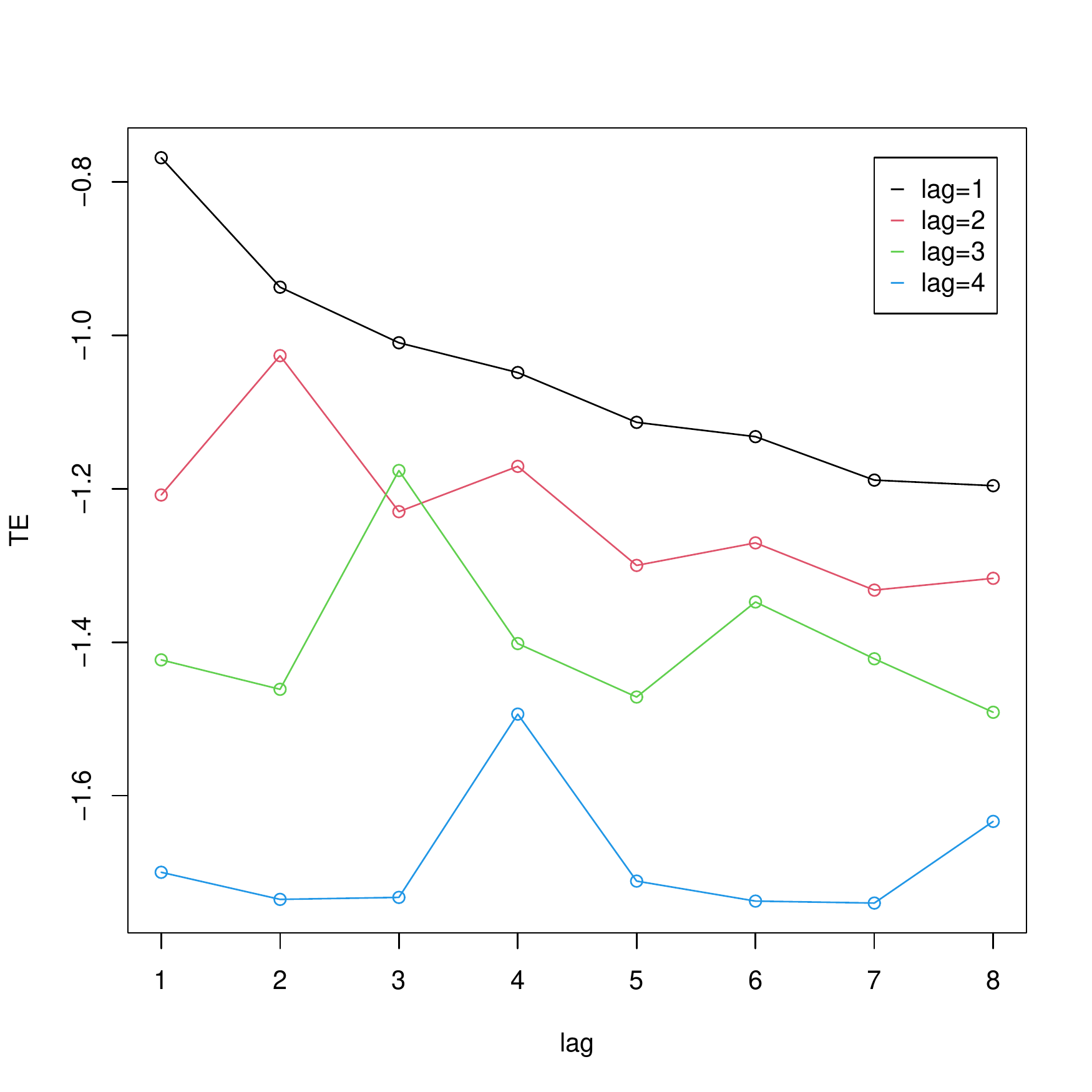}
	\caption{Results of the simulation experiment 4.}
	\label{fig:sim2result}
\end{figure}

The results of the simulation experiments are shown in Figure \ref{fig:sim}-\ref{fig:sim2result}. One of the simulated time series data of the first system is shown in Figure \ref{fig:sim1data1}. The estimated TE values from the state variable to the output variable of the system with four time lags are shown in Figure \ref{fig:sim1result1}. It can be learned that the maximum of the TE values correspond to the time lags of the system in all the four cases. 

One of the simulated time series data of the second system is shown in Figure \ref{fig:sim1data2}, from which it can be easily learned that the output of the system follows the state of the system with the pre-specified time lag. The estimated TE value from the state variable to the output variable of the system with four time lags are shown in Figure \ref{fig:sim1result2}. It can be learned from the Figure that the maximum of the TE values corresponds to the time lags of the system in all the four cases. The difference between the results of the first and second systems is that the TE value of the second system does not return to 0 with time lags as that of the first system does. This is due to the dynamics introduce by the sine term of the second system.

The results of the simulation experiment 3 are shown in Figure \ref{fig:sim1data3} and \ref{fig:sim1result3}. The simulate data are illustrated in Figure \ref{fig:sim1data3}, which shows the stochastic dynamics of the simulation system. Since the system is first-order, the dynamics of it is much complex that the previous two ones. The estimated TE values from the state variable of the system to the output variable of the system with respect to four time lags are shown in Figure \ref{fig:sim1result3}. It can be learned from it that the four time lag are all correctly identified by the maximum of TE values in all the four cases. The estimated TE values in four cases are all increasing to their maximum and then vanishing gradually as time lag increases.

The simulated data and the estimated results of the simulation experiment 4 are shown in Figure \ref{fig:sim2} and \ref{fig:sim2result} respectively. It can be learned from Figure \ref{fig:sim2result} that our method can identify the four time lag in the simulated systems through the maximum of the estimated TE values correctly.

\section{Real data}
\label{sec6}
We also demonstrated our method with a real data on power consumption of the Tetouan city in the UCI machine learning repository \cite{Asuncion2007}. The dataset is the historical data on power consumption of three different distribution networks (Quads, Smir and Boussafou) of the Tetouan city in north Morocco in 2017 \cite{Salam2018}. The aim of the dataset is to predict power consumption with weather factors. The data of the three networks are different due to seasonal and social reasons. Besides the time series data of power consumption of three networks, five weather factors that affect the power consumption, including temperature, humidity, wind speed, general diffuse flows, diffuse flows, are also contained in the dataset. The sample rate of the data is 10 minutes per sample. 

It is well-known that the change of weather will affect the power consumption after a peroid of time. So, we want to study the weather's effect on power consumption in terms of time lag. We will apply our method to the data to study the time lag from weather to power consumption. In the experiment, the data of five working days (2017-10-23$\sim$2017-10-27) were used since there is different between working day and holiday. We estimated the TE from five weather factors to the power consumption of the three networks. The time lag is from 1 hour to 24 hours because we are interested in how the estimated TE, i.e., the causality of weather's effect on power consumption, change in a whole day.

\begin{figure}
	\centering
	\includegraphics[width=\textwidth]{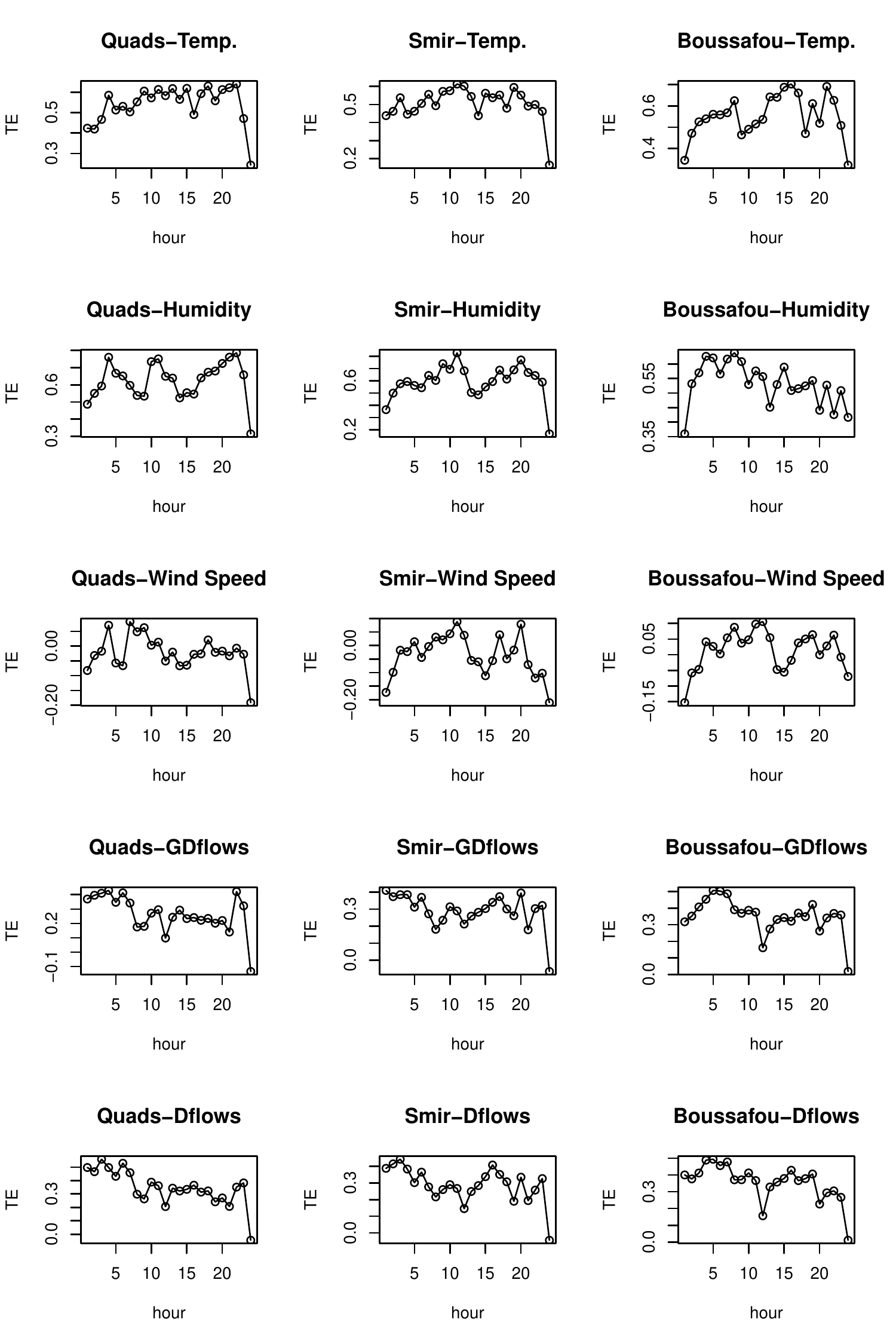}
	\caption{Results of the experiment with the real data of the Tetouan city.}
	\label{fig:tetouan1}
\end{figure}

The estimation results of the experiment is shown in Figure \ref{fig:tetouan1}, in which each row is for a weather factor and each column is for a network. Each plot presents the change of TE from 1 hour to 24 hours time lag. It can be learned from it that 1) there is a time lag (about 4 hours) in almost all the estimated TEs from weather factors to the power consumption of the three networks, during which the causal effect became stronger quickly; 2) the time lag from temperature to power consumption when the causal effect reach its maximum is much longer and slower; 3) the estimated TE from the other four factors to power consumption of the three networks can be considered as having two peaks in each half day, which means there are two time lags due to the daily pattern of weather factors.  

\section{Conclusions}
\label{sec7}
In this paper, we propose to use the CE-based TE estimator for identifying time lag in dynamical systems. We use both simulated and real data to test our method in the experiments. Four stochastic dynamical systems are designed to simulate time series data with time lags. Experimental results show that our method can identify the time lags in these systems from the simulated data. A real data of power consumption of the Tetouan city is used to demonstrate the effectiveness of our method. The estimated results show that our method can identify the pattern of time lags from weather factors to power consumption of the Tetouan city. 

\appendix
\section{Code}
The codes are available at \url{https://github.com/majianthu/timelag}.

\bibliographystyle{unsrt}
\bibliography{lag}

\begin{thebibliography}{10}

\bibitem{Bjorklund2003}
S.~Bjorklund and L.~Ljung.
\newblock A review of time-delay estimation techniques.
\newblock In {\em 42nd IEEE International Conference on Decision and Control
  (IEEE Cat. No.03CH37475)}, volume~3, pages 2502--2507 Vol.3, Dec 2003.

\bibitem{Knapp1976}
C.~Knapp and G.~Carter.
\newblock The generalized correlation method for estimation of time delay.
\newblock {\em IEEE Transactions on Acoustics, Speech, and Signal Processing},
  24(4):320--327, August 1976.

\bibitem{Gu2011}
Guojun Gu and Robert~F. Adler.
\newblock Precipitation and temperature variations on the interannual time
  scale: Assessing the impact of {ENSO} and volcanic eruptions.
\newblock {\em Journal of Climate}, 24(9):2258 -- 2270, 2011.

\bibitem{Kiviet2014}
Daniel~J Kiviet, Philippe Nghe, Noreen Walker, Sarah Boulineau, Vanda
  Sunderlikova, and Sander~J Tans.
\newblock Stochasticity of metabolism and growth at the single-cell level.
\newblock {\em Nature}, 514(7522):376--379, 2014.

\bibitem{Mars1982}
N.J.I. Mars and G.W. {van Arragon}.
\newblock Time delay estimation in non-linear systems using average amount of
  mutual information analysis.
\newblock {\em Signal Processing}, 4(2):139--153, 1982.

\bibitem{Wen2009}
Fei Wen and Qun Wan.
\newblock Time delay estimation based on mutual information estimation.
\newblock In {\em 2009 2nd International Congress on Image and Signal
  Processing}, pages 1--5, Oct 2009.

\bibitem{Albers2012}
D.~J. Albers and George Hripcsak.
\newblock Using time-delayed mutual information to discover and interpret
  temporal correlation structure in complex populations.
\newblock {\em Chaos: An Interdisciplinary Journal of Nonlinear Science},
  22(1):013111, 2012.

\bibitem{Wilmer2012}
Andreas Wilmer, Marc de~Lussanet, and Markus Lappe.
\newblock Time-delayed mutual information of the phase as a measure of
  functional connectivity.
\newblock {\em {PLoS ONE}}, 7(9):e44633, 2012.

\bibitem{infobook}
Thomas~M Cover and Joy~A Thomas.
\newblock {\em Elements of information theory}.
\newblock John Wiley \& Sons, 2012.

\bibitem{Schreiber2000}
Thomas Schreiber.
\newblock Measuring information transfer.
\newblock {\em Physical Review Letters}, 85:461--464, Jul 2000.

\bibitem{Granger1969}
Clive~WJ Granger.
\newblock Investigating causal relations by econometric models and
  cross-spectral methods.
\newblock {\em Econometrica: Journal of the Econometric Society}, pages
  424--438, 1969.

\bibitem{Granger1980}
Clive~WJ Granger.
\newblock Testing for causality: a personal viewpoint.
\newblock {\em Journal of Economic Dynamics and Control}, 2:329--352, 1980.

\bibitem{Lahiri2017}
Sourabh Lahiri, Philippe Nghe, Sander~J Tans, Martin~Luc Rosinberg, and David
  Lacoste.
\newblock Information-theoretic analysis of the directional influence between
  cellular processes.
\newblock {\em PLoS ONE}, 12(11):e0187431, 2017.

\bibitem{Rosinberg2018}
M.~L. Rosinberg, G.~Tarjus, and T.~Munakata.
\newblock Influence of time delay on information exchanges between coupled
  linear stochastic systems.
\newblock {\em Physical Review E}, 98:032130, Sep 2018.

\bibitem{Marzbanrad2015}
Faezeh Marzbanrad, Yoshitaka Kimura, Marimuthu Palaniswami, and Ahsan~H
  Khandoker.
\newblock Quantifying the interactions between maternal and fetal heart rates
  by transfer entropy.
\newblock {\em PloS ONE}, 10(12):e0145672, 2015.

\bibitem{Oh2023}
YongKyung Oh, JiIn Kwak, and Sungil Kim.
\newblock Time delay estimation of traffic congestion propagation due to
  accidents based on statistical causality.
\newblock {\em Electronic Research Archive}, 31(2):691--707, 2023.

\bibitem{faes2013compensated}
Luca Faes, Giandomenico Nollo, and Alberto Porta.
\newblock Compensated transfer entropy as a tool for reliably estimating
  information transfer in physiological time series.
\newblock {\em Entropy}, 15(1):198--219, 2013.

\bibitem{Vicente2011}
Raul Vicente, Michael Wibral, Michael Lindner, and Gordon Pipa.
\newblock Transfer entropy—a model-free measure of effective connectivity for
  the neurosciences.
\newblock {\em Journal of Computational Neuroscience}, 30(1):45--67, 2011.

\bibitem{Ma2011}
Jian {Ma} and Zengqi {Sun}.
\newblock Mutual information is copula entropy.
\newblock {\em Tsinghua Science \& Technology}, 16(1):51--54, 2011.
\newblock See also arXiv preprint arXiv:0808.0845, 2008.

\bibitem{ma2021estimating}
Jian Ma.
\newblock Estimating transfer entropy via copula entropy.
\newblock {\em arXiv preprint arXiv:1910.04375}, 2019.

\bibitem{Liu2022}
Hao Liu, Dechang Pi, Shuyuan Qiu, Xixuan Wang, and Chang Guo.
\newblock Data-driven identification model for associated fault propagation
  path.
\newblock {\em Measurement}, 188:110628, 2022.

\bibitem{Han2022}
Muye Han and Jinsheng Zhou.
\newblock Multi-scale characteristics of investor sentiment transmission based
  on wavelet, transfer entropy and network analysis.
\newblock {\em Entropy}, 24(12), 2022.

\bibitem{Hu2022}
Wei Hu, Qiuting Guo, Wei Wang, Weiheng Wang, and Shuhong Song.
\newblock Research on user loss contribution calculation of high-loss
  distribution area based on transfer entropy.
\newblock In {\em 2022 China International Conference on Electricity
  Distribution (CICED)}, pages 499--502, 2022.

\bibitem{Amornbunchornvej2021}
Chainarong Amornbunchornvej, Elena Zheleva, and Tanya Berger-Wolf.
\newblock Variable-lag {Granger} causality and transfer entropy for time series
  analysis.
\newblock {\em ACM Transactions on Knowledge Discovery from Data},
  15(4):1–30, Jun 2021.

\bibitem{Wing2016}
Simon Wing, Jay~R. Johnson, Enrico Camporeale, and Geoffrey~D. Reeves.
\newblock Information theoretical approach to discovering solar wind drivers of
  the outer radiation belt.
\newblock {\em Journal of Geophysical Research: Space Physics},
  121(10):9378--9399, 2016.

\bibitem{Staniek2008}
Matth\"aus Staniek and Klaus Lehnertz.
\newblock Symbolic transfer entropy.
\newblock {\em Physical Review Letters}, 100:158101, Apr 2008.

\bibitem{Wang2017}
Xuguang Wang, Wei Dong, and Yan Zhang.
\newblock Copula based time delay estimation for linear first order system.
\newblock In {\em Proceedings of the 2017 International Conference on Applied
  Mathematics, Modelling and Statistics Application (AMMSA 2017)}, pages
  414--419. Atlantis Press, 2017.

\bibitem{nelsen2007}
Roger~B Nelsen.
\newblock {\em An introduction to copulas}.
\newblock Springer Science \& Business Media, 2007.

\bibitem{joe2014}
Harry Joe.
\newblock {\em Dependence modeling with copulas}.
\newblock CRC press, 2014.

\bibitem{sklar1959fonctions}
M~Sklar.
\newblock Fonctions de repartition an dimensions et leurs marges.
\newblock {\em Publ. inst. statist. univ. Paris}, 8:229--231, 1959.

\bibitem{Kraskov2004}
Alexander Kraskov, Harald St{\"o}gbauer, and Peter Grassberger.
\newblock Estimating mutual information.
\newblock {\em Physical Review E}, 69(6):066138, 2004.

\bibitem{Asuncion2007}
Arthur Asuncion and David Newman.
\newblock {UCI} machine learning repository, 2007.

\bibitem{Salam2018}
Abdulwahed Salam and Abdelaaziz~El Hibaoui.
\newblock Comparison of machine learning algorithms for the power consumption
  prediction : Case study of {Tetouan} city.
\newblock In {\em 2018 6th International Renewable and Sustainable Energy
  Conference (IRSEC)}, pages 1--5, 2018.

\end{thebibliography}

\end{document}